\documentclass[letterpaper, 10 pt, journal, twoside]{IEEEtran}

\usepackage{cite}
\usepackage{amsmath,amssymb,amsfonts}
\usepackage{algorithmic}
\usepackage{graphicx}
\usepackage{textcomp}
\usepackage{xcolor}
\usepackage{algorithm}
\usepackage{algorithmic}
\usepackage{amsthm}
\usepackage{multirow}
\usepackage{booktabs}
\usepackage{stfloats}
\usepackage{subcaption}  

\newtheoremstyle{plain_noitalic}  
  {\topsep}                       
  {\topsep}                       
  {\normalfont}                   
  {}                              
  {\bfseries}                     
  {.}                             
  {.5em}                          
  {}                              

\theoremstyle{plain_noitalic}
\newtheorem{theorem}{Theorem}
\newtheorem{lemma}{Lemma}

\newtheorem{assumption}{Assumption}

\theoremstyle{definition}
\newtheorem{definition}{Definition}

\begin{document}
%
\title{Collaborative Task Allocation for Heterogeneous Multi-Robot Systems Through Iterative Clustering}

\author{David R. Martin$^{1}$, Brooks A. Butler$^{1}$, Scott Nivison$^{2}$, Magnus Egerstedt$^{1}$, Mohammad Abdullah Al Faruque$^{1}$, and Pramod Khargonekar$^{1}$
\thanks{Manuscript received: July, 7, 2025; Revised September, 27, 2025; Accepted October, 22, 2025. This paper was recommended for publication by Editor Chao-Bo Yan upon evaluation of the Associate Editor and Reviewers' comments.
This work is sponsored in part by the Air Force Research Laboratory, Munitions Directorate (RWTA), Eglin AFB, FL. Opinions, findings and conclusions, or recommendations are those of the authors and do not necessarily reflect the views of the sponsoring agencies.} 
\thanks{$^{1}$David R. Martin ({\tt\footnotesize davidrm3@uci.edu}), Brooks A. Butler, Magnus Egerstedt, Mohammad Abdullah Al Faruque ({\tt\footnotesize alfaruqu@uci.edu}), and Pramod Khargonekar ({\tt\footnotesize pramod.khargonekar@uci.edu}) are with the Henry Samueli School of Engineering, Department of Electrical Engineering and Computer Science, University of California, Irvine.}%
\thanks{$^{2} $Scott Nivison is with the Air Force Research Laboratory, Munitions Directorate, Eglin AFB, FL.}

\thanks{Digital Object Identifier (DOI): 10.1109/LRA.2025.3632086}

\thanks{
\textcopyright{} 2025 IEEE. Personal use of this material is permitted.  Permission from IEEE must be obtained for all other uses, in any current or future media, including reprinting/republishing this material for advertising or promotional purposes, creating new collective works, for resale or redistribution to servers or lists, or reuse of any copyrighted component of this work in other works.
}
}

\markboth{IEEE Robotics and Automation Letters. Preprint Version. Accepted October, 2025}
{Martin \MakeLowercase{\textit{et al.}}: Collaborative Task Allocation for Heterogeneous Multi-Robot Systems Through Iterative Clustering} 

\maketitle

\begin{abstract} Multi-robot systems face the challenge of efficiently allocating teams of heterogeneous robots to tasks. The task allocation problem is complicated by collaborative interactions between robots where teams of robots develop emergent capabilities that enable them to complete tasks that would be inefficient or impossible for individual robots. To address these challenges, we present an iterative clustering algorithm for collaborative task allocation in heterogeneous multi-robot systems. This approach partitions the computationally intractable global optimization problem into smaller, tractable subproblems by iteratively forming clusters of robots and tasks, then optimizing assignments within each cluster. By ensuring robots remain clustered with their currently assigned tasks, we guarantee monotonic improvement in overall utility with each iteration. We analyze the convergence of the algorithm and characterize how cluster size constraints determine which suboptimal assignments could trap the algorithm. In simulation, iterative clustering consistently outperforms simulated annealing, and a group-based auction in both computation time and solution quality, and outperforms a hedonic game approach in solution quality.
\end{abstract}

\begin{IEEEkeywords}
Multi-robot systems, task planning, multi-robot task allocation (MRTA), cooperating robots
\end{IEEEkeywords}

%
\IEEEpeerreviewmaketitle

\section{Introduction}

\IEEEPARstart{M}{ulti-robot systems} have demonstrated significant potential in numerous domains, including disaster relief \cite{ghassemi2022multi}, sensor fusion \cite{hu2023collaboration}, construction \cite{dai2024dynamic}, warehousing \cite{liu2022multirobot,shan2024distributed}, and beyond \cite{athira2024systematic,chakraa2023optimization}. Of particular interest are robot teams where collaboration enables new capabilities, allowing them to complete tasks that would be impossible or highly inefficient for individual robots. Recent examples highlight this synergistic potential: in \cite{hu2023collaboration}, multiple drones equipped with cameras collaboratively estimate 3D object depth with higher accuracy than lidar, in \cite{nguyen2023mutualistic}, aquatic robots transport terrestrial robots across water bodies to reach otherwise inaccessible goals, in \cite{dai2024dynamic} and \cite{shan2024distributed}, robot teams collaborate to carry oversized payloads.

A challenge in collaborative multi-robot systems is the multi-robot task allocation (MRTA) problem, which is the process of determining optimal robot-to-task assignments that maximize system performance \cite{chakraa2023optimization}. Researchers have addressed MRTA using diverse approaches, including the Hungarian algorithm \cite{chopra2017distributed}, mixed integer programming \cite{notomista2022}, auctions \cite{shan2024distributed, Bai9781631}, game-theoretic approaches \cite{zhang2024coalition,zhang2024distributed}, metaheuristics (such as simulated annealing \cite{junwei2014study} and evolutionary algorithms \cite{younas2018efficient,guo2024effective}), clustering techniques \cite{you2023novel,martin2023multi}, and machine learning methods \cite{dai2024dynamic,goarin2024graph}. Distributed algorithms such as group-based auctions \cite{Bai9781631} and game theoretic approaches \cite{zhang2024distributed} have shown a remarkable ability to scale efficiently to large systems.

Despite recent advances, existing algorithms struggle to account for collaborative synergy in multi-robot teams. Most existing approaches assume predetermined team sizes and capability requirements for each task \cite{dai2024dynamic,notomista2022,younas2018efficient}. However, in many systems, the optimal team size may be unclear. When resources are abundant and tasks are few, it may be advantageous to assign more resources to each task. Conversely, when resources are constrained, it may be necessary to reduce team sizes or omit less important tasks completely. Collaborative dynamics complicate resource allocation. Existing works that allow variable team sizes generally rely on assigning individual robots to tasks where they provide the greatest marginal contribution \cite{zhang2024coalition, karam2025resource}. However, these approaches struggle in situations where robots' contributions are not easily separable from those of their teammates. For instance, a task requiring three robots might yield no utility until all three are assigned. In general, these methods fail in scenarios requiring simultaneous multi-robot reassignments, where a single reassignment provides no benefit.

To account for collaborative team effects, we model the MRTA problem as a coalition structure generation problem \cite{shehory1998methods}, partitioning robots into coalitions assigned to each task with the goal of optimizing total utility. While coalition structure generation is NP-hard \cite{rahwan2015coalition}, we propose a novel iterative clustering algorithm that scales effectively. The proposed method directly addresses the key limitations of existing methods by: (1) providing a computationally efficient solution to the coalition structure generation problem, (2) modeling team utility as a nonlinear function that can be greater than the sum of the individual robot contributions, and (3) determining appropriate team compositions for each task while accounting for this collaborative synergy.

The proposed method iteratively forms clusters of robots and tasks, then optimizes assignments within each cluster. We develop two variants of this approach: one that forms clusters randomly and another that uses heuristics to guide cluster formation. We prove that assignment utility increases monotonically with each iteration and analyze how cluster size constraints affect the algorithm’s ability to escape suboptimal solutions. In simulation, the iterative clustering approach achieves 99.9\% of optimal utility for small problems, outperforms exisiting methods for large problems, and exhibits fast convergence that scales effectively to problems with up to 1000 robots and 500 tasks.
\section{System Model and Problem Formulation}
This work focuses on ST-MR-IA \cite{gerkey2004formal} problems where ST indicates single-task robots (each robot can perform only one task at a time), MR denotes multi-robot tasks (tasks may be performed by multiple robots), and IA refers to instantaneous assignment (planning without considering future allocations).

\subsection{System Model} \label{section:SysModel}

Consider a set of $\nu$ robots $\mathcal{R} = \{r_1,r_2, \dots, r_\nu \}$ to be assigned to a set of $\mu$ tasks $\mathcal{T} = \{t_1,t_2,\dots, t_\mu \}$. 

\subsubsection{Assignments} The individual assignment of each robot $r_i$ is represented as an integer $a_i \in \{0,1,\dots,\mu\}$ such that $a_i=0$ means $r_i$ is unassigned and $a_i=j$ for $j \neq 0$ means $r_i$ is assigned to task $t_j$. A global assignment $A = \{S_0, S_1,\dots,S_{\mu}\}$ partitions the set of all robots $\mathcal{R}$ into $\mu+1$ teams $S_0,S_1,...,S_\mu \subseteq \mathcal{R}$ where team $S_j = \{r_i \in \mathcal{R}: a_i = j\}$ contains the robots assigned to task $t_j$ for $j \neq 0$ and $S_0 = \{r_i \in \mathcal{R}: a_i = 0\}$ contains all unassigned robots. Because each robot is assigned to no more than one task, an assignment $A = \{S_0, S_1, \dots,S_\mu \}$ forms  a partition of $\mathcal{R}$ meaning  $S_\alpha \cap S_\beta = \emptyset$ for $\alpha \neq \beta$ and  $\bigcup_{j=0}^{\mu}S_j = \mathcal{R}$. Let $\mathcal{A}$ denote the set of all possible assignments.

\subsubsection{Robot Types}
Each robot $r_i \in \mathcal{R}$ is characterized by a single robot type that represents different hardware or software capabilities (e.g. UAV, rover... etc.). Robot type for robot $r_i$ is denoted $z_i \in \{1,2,...\kappa\}$, where $\kappa$ is the total number of distinct robot types. Robots of the same type are considered identical in terms of their capabilities.

\subsubsection{Teams}
Each robot team $S_j$ has an associated team composition vector $C_j = [c_1,c_2,\dots,c_\kappa]$ where $c_\lambda$ denotes the number of robots of type $\lambda$ in the team. While the team size is variable, we constrain the maximum team size to an integer limit $L$: $|S_j| \leq L$ for $j \neq 0$. The number of unassigned robots $|S_0|$ is unconstrained.
\subsubsection{Team Utility Modeling}  \label{sec_util_modeling}
Each task $t_j \in \mathcal{T}$ is associated with a benefit function 
$B_j(C): \mathbb{N}_0^\kappa \rightarrow \mathbb{R}$, which specifies the 
utility of assigning a team of robots with team composition vector 
$C = [c_1, \dots, c_\kappa]$ to task $t_j$. Intuitively, it captures how 
good a particular team is for completing the task. For example, consider a search 
and rescue mission with two robot types ($\kappa = 2$): drones and ground vehicles. A single drone may be unable to complete the rescue and 
provide no benefit ($B_j([1,0]) = 0$). A team with one drone for victim 
localization and one ground vehicle for extraction could provide high 
utility ($B_j([1,1]) = 100$), while adding another drone may provide quicker localization ($B_j([2,1]) = 120$). Benefit functions thus 
provide a quantitative model to directly represent complex team interactions, including collaborative synergies and marginally decreasing returns. We do not impose any particular structure or form on the benefit functions since they depend on the specifics of tasks, robots, and mission priorities. Within the task assignment problem, benefit functions are treated as known inputs, under the assumption that the system designer can specify or evaluate them using domain expertise or data-driven methods.
\begin{assumption}
We assume that the benefit functions of all tasks, robot capabilities, and locations of robots and tasks are fully known and deterministic.
\end{assumption}

\subsubsection{Team Formation Cost} Each task $t_j$ has a team formation cost function $J_j(S_j): 2^{\mathcal{|R|}} \rightarrow \mathbb{R}$ which quantifies the cost of assembling the team $S_j$ prior to task initiation. This cost may encompass factors such as travel time and energy, as well as any time spent waiting for team members to arrive. Like the benefit function, the team formation cost function depends  on the nature of the task and the assigned robots. For the simulated problems, each $J_j(S_j)$ is the cumulative distance traveled by robots to the task location:

\begin{equation}\label{costFunction}
    J_j(S_j) = \sum_{r_i \in S_j}\sqrt{(x_i-x_j)^2+(y_i-y_j)^2}
\end{equation} where $(x_i, y_i)$ is the location of robot $r_i$ and $(x_j, y_j)$ is the location of task $t_j$.

\subsubsection{Utility}

The net utility of completing a task $t_j$ with team $S_j$ is the benefit subtracted by the team formation cost:
\begin{equation} \label{taskUtility}
U(S_j) = {B_{j}(C_j)} - J_j(S_j).
\end{equation}

The utility of the unassigned robots is defined to be zero: $U(S_0) = 0$. Define the  function $\mathcal{U}(A)$ to be be the total utility produced by an assignment $A$:

\begin{equation}
\mathcal{U}(A) = \sum_{S_j \in A} U(S_j)
\end{equation}

\subsection{Problem Formulation}

The objective of collaborative task assignment is to maximize the total utility generated across all teams. We include a user-defined team size limit $L$ which represents the largest reasonable number of robots for any single task.

\begin{equation}
\label{objective}
\begin{aligned}
   & \max_{A\in \mathcal{A}} \sum_{S_j \in A} U(S_j) \\
   & \text{subject to} \quad |S_j| \leq L, \quad \forall j \in \{1,2,\ldots,\mu\}
\end{aligned}
\end{equation}

This is a special case of the coalition structure generation problem discussed in \cite{rahwan2015coalition}, where each team $S_j \in A$ has its own utility function associated with a specific task. As discussed in \cite{rahwan2015coalition}, coalition formation is an NP-hard problem, making it computationally challenging even under restrictive assumptions. The proposed iterative clustering approach provides a computationally efficient method for obtaining quick, anytime solutions to this otherwise intractable problem.

\section{Methodology}

\begin{figure*}[t]  
    \centering
    \includegraphics[width=.8\textwidth]{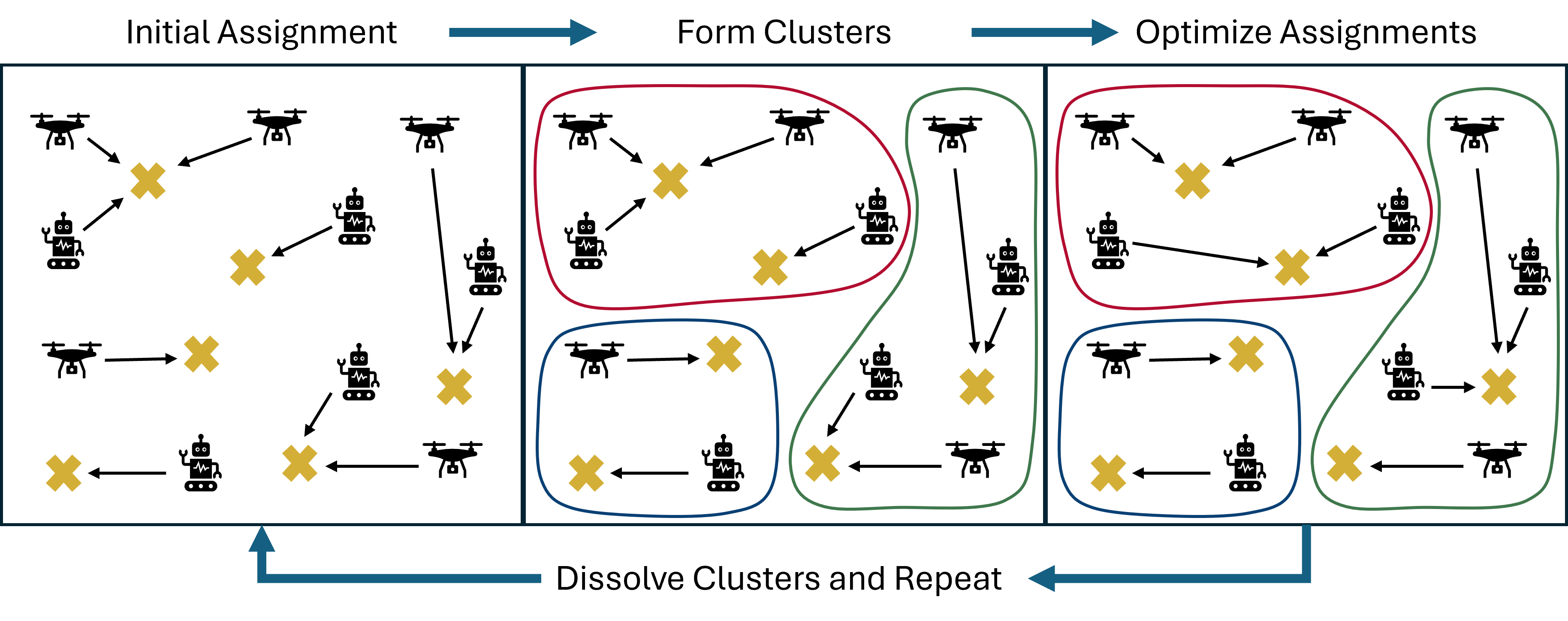}
    \caption{Illustration of one iteration of Algorithm~\ref{alg:mainAlgorithm}. First, robots and tasks are partitioned into clusters without separating robots from their currently assigned task. Then, an optimal assignment algorithm optimizes the assignments within each cluster. The clusters are then dissolved and the process is repeated.}
    \label{fig:yourlabel}
\end{figure*}

The task allocation problem is addressed using an iterative approach where each iteration has two phases. In the first phase, robots and tasks are split into a set of clusters $G = \{G_1,\ldots, G_{\ell}\}$, where the number of clusters $\ell$ may vary. Each cluster  $G_k= (G_k^r,  G_k^t)$ is defined by a subset of robots $G_k^r \subseteq \mathcal{R}$ and a subset of tasks $G_k^t \subseteq \mathcal{T}$. Clustering in this way partitions the large, computationally intractable problem into a set of smaller, tractable subproblems. In the second phase, an optimal assignment algorithm is used to compute the optimal assignments within the clusters i.e., $A^*_k = \{S_0^k\} \cup \{S_j:t_j \in G_k^t\}$ for all $k=1,2,\ldots,\ell$ where $S_j \subseteq G_k^r$ is the set of robots assigned to task $t_j \in G_k^t$ and $S_0^k \subseteq G_k^r$ includes any robots not assigned to a task. Recall our problem formulation does not require all robots to be assigned to tasks. We construct the global assignment by combining cluster assignments, with all unassigned robots merged into $S_0$:
\begin{equation} \label{UnionDef}
A = \overline{\bigcup}_{k=1}^{\ell}A^*_k := \{S_0 = \bigcup_{k=1}^{\ell}S_0^k\} \cup \bigcup_{k=1}^{\ell}\{S_j\in A_k^*: j \neq 0\}.
\end{equation}

While a single iteration produces optimal assignments within each cluster, it does not ensure global optimality. However, by repeatedly re-clustering the problem while ensuring robots remain clustered with their currently assigned tasks, we show that the global assignment progressively improves. Algorithm~\ref{alg:mainAlgorithm} outlines our overall solution framework. Each choice of cluster formation algorithm and optimal assignment algorithm leads to a concrete and distinct algorithm for solving the main problem.

\begin{algorithm}[h]
\caption{Task Assignment Through Iterative Clustering}\label{alg:mainAlgorithm}
\begin{algorithmic}
\STATE \textbf{Input:} Robots $\mathcal{R}$, Tasks $\mathcal{T}$, Number of iterations. 
\STATE \textbf{Output:} Assignment $A = \{S_0, S_1, \dots, S_\mu \}$ 
\STATE 
\STATE Initialize $A = \{ \mathcal{R}, \emptyset, \dots, \emptyset \}$ with all robots unassigned;
\FOR{each iteration}
    \STATE $G \gets \text{ClusterFormation}(A, \mathcal{R}, \mathcal{T})$; \textit{(Algorithm~\ref{alg:Clustering})}
    \FOR{each cluster $G_k \in G$}
        \STATE $A^*_k \gets \text{OptimalAssignment}(G_k)$; \textit{(Section \ref{sec:clusterOptimization})}
    \ENDFOR
    \STATE $A \gets \overline{\bigcup}_{k=1}^{\ell} A^*_k$; \textit{(Form global assignment per Eq.~\ref{UnionDef})}
\ENDFOR
\STATE Return $A$
\end{algorithmic}
\end{algorithm}


\subsection{Cluster Formation}
The cluster formation algorithm builds clusters of robots and tasks, limiting each cluster to user-defined maximums of $L_r$ robots and $L_t$ tasks while ensuring robots are clustered with their currently assigned tasks. Cluster size constraints present a trade-off: increasing $L_t$ and $L_r$ reduces the number of suboptimal assignments that could trap Algorithm~\ref{alg:mainAlgorithm}, but also increases the computational complexity of the task assignment problem within each cluster.

\begin{definition} \label{def:feasibleCluster}
A cluster $G_k = (G_k^r,G_k^t)$ is feasible if it satisfies the following constraints:
\begin{subequations}\label{clusterConstraints}
\begin{align}
    |G_k^r| &\leq L_r \label{eq:a}\\
    |G_k^t| &\leq L_t \label{eq:b}\\
    S_j &\subseteq G_k^r \quad \forall j : t_j \in G_k^t \label{eq:c}
\end{align}
\end{subequations}
\end{definition}
where constraints \ref{eq:a} and \ref{eq:b} limit the number of robots and tasks in the cluster and \ref{eq:c} guarantees robots are clustered with their currently assigned task.

Any clustering algorithm can be employed that ensures all clusters are feasible. Ensuring robots remain clustered with their currently assigned tasks guarantees that the utility of the assignment is monotonically increasing with each iteration.


\begin{lemma}\label{nondecreasingLemma}
Let $A^i$ and $A^{i+1}$ denote the assignments before and after the $i$-th iteration of Algorithm~\ref{alg:mainAlgorithm}, respectively. If clusters are feasible and assignments within each cluster are optimal, then $\mathcal{U}(A^i) \leq \mathcal{U}(A^{i+1})$.
\end{lemma}

\begin{proof} Let $G = \{G_1...G_{\ell}\}$ be a feasible clustering. Since robots are clustered with their currently assigned tasks, the assignment $A^i$ can be written as the combination of the assignments within each cluster: $A^i = \overline{\bigcup}_{k=1}^{\ell} A^i_k$ (Eq. \ref{UnionDef}). The value of the assignment $A^i$ is the sum of the utilities generated within each cluster.

\begin{equation}
    \mathcal{U}(A^i) = 
    \sum_{k=1}^{\ell}\mathcal{U}(A^i_k) =
    \sum_{k=1}^{\ell}\sum_{S_j \in A^i_k}U(S_j)
\end{equation}

Optimal assignments $A_k^*$ are made within each cluster $G_k$. Since $\mathcal{U}(A^i_k) \leq \mathcal{U}(A^*_k)$ it follows that after one iteration:

$\mathcal{U}(A^i) = 
\sum_{k=1}^{\ell} \mathcal{U}(A^i_k) \leq
\sum_{k=1}^{\ell} \mathcal{U}(A^*_k) = 
\mathcal{U}(A^{i+1})$.
\end{proof}

The cluster formation algorithm (Algorithm~\ref{alg:Clustering}) begins by creating initial clusters by grouping each task with the robots assigned to it, while any unassigned task or robot starts as its own cluster. The algorithm then keeps track of two lists: incomplete clusters ($G^i$), which can still be enlarged, and complete clusters ($G^c$), which cannot grow further without violating the robot limit $L_r$ or the task limit $L_t$. At each step, the algorithm randomly picks an incomplete cluster $G_\alpha$ and searches for other clusters that can be merged with it without exceeding the size limits. If such candidates exist, one is chosen and merged with $G_\alpha$; if none exist, $G_\alpha$ is moved to the complete clusters list. This process repeats until every cluster is complete, meaning no further merges are possible within the given limits.


\begin{algorithm}[h]
\caption{Clustering Algorithm}\label{alg:Clustering}
\begin{algorithmic}
\STATE \textbf{Input:} Assignment $A$, Robots $\mathcal{R}$, Tasks $\mathcal{T}$, Limits $L_r, L_t$.
\STATE \textbf{Output:} Clustering $G = \{G_1...G_{\ell}\}$ 
\STATE 
\STATE Create initial incomplete clusters $G^i$ by grouping tasks
\STATE \ with their assigned robots. Any unassigned robot or task
\STATE \ will form its own cluster;
\STATE Initialize empty list of complete clusters: $G^c \gets  \emptyset$;
\WHILE{$G^i \neq \emptyset$}
    \STATE Select an incomplete cluster: $G_\alpha \sim \text{Uniform}(G^i)$;
        \STATE Find merge candidates: $M \gets \{G_\gamma \in G^i :$
        \STATE $\quad\quad |G_\alpha^r| + |G_\gamma^r| \leq L_r, |G_\alpha^t| + |G_\gamma^t| \leq L_t, \alpha \neq \gamma\}$;
        \IF{$M = \emptyset$}
            \STATE Add $G_\alpha $ to complete clusters list: $G^c \gets G^c \cup G_\alpha$;
            \STATE Remove from incomplete clusters list: $G^i \gets G^i \setminus G_\alpha$;
        \ELSE
            \STATE Choose a cluster to merge with:
            \STATE \ $G_\beta \sim $ MergePolicy$(M,G_\alpha)$; \textit{(Section \ref{sec:mergeSelectionSection})}
            \STATE Merge clusters: $G^i \gets G^i \setminus \{G_\alpha,G_\beta\}\cup (G_\alpha \cup G_\beta)$;
        \ENDIF
\ENDWHILE
\STATE \text{Return } $G^c$
\end{algorithmic}
\end{algorithm}


This work considers two merge policies for selecting which clusters to merge. The first is a uniform random selection of all feasible merge candidates. The second is a heuristic-based selection that forms higher-quality clusters by accounting for robot capabilities and spatial locations.

\subsection{Merge Candidate Selection Heuristics} \label{sec:mergeSelectionSection}

This section develops heuristics that guide cluster formation by aligning robot capabilities with task requirements while promoting spatial proximity.

\subsubsection{Capability Matching }

Robots should be clustered with tasks for which their capabilities are well suited. Therefore, clusters $G_\alpha$ and $G_\beta$ are only merged if at least one robot $r_i \in G_\alpha$ can contribute to a task $t_j \in G_\beta$ or vice versa. A robot $r_i$ is considered to be capable of contributing to a task $t_j$ if there exists a team composition vector $C$ with at least one robot of type $z_i$ such that $B_j(C) > 0$. Define a symmetric capability matching function $m(G_\alpha, G_\beta)$:

\begin{equation} \label{capMatch}
    m(G_\alpha, G_\beta) = 
    \begin{cases}
        1 & \text{if } \exists r_i \in G_\alpha^r, t_j \in G_\beta^t, C \in \mathbb{N}_{0}^{\kappa}: \\
          & \quad B_j(C) > 0 \text{ and } C_{[z_i]} > 0 \\
        1 & \text{if } \exists r_i \in G_\beta^r, t_j \in G_\alpha^t, C \in \mathbb{N}_{0}^{\kappa}: \\
          & \quad B_j(C) > 0 \text{ and } C_{[z_i]} > 0 \\
        0 & \text{otherwise}
    \end{cases}
\end{equation}

\subsubsection{Distance-Based Cluster Merging Probability}

To minimize travel distance, we prioritize merging clusters that are closer together. Define a cluster $G_k$'s center of mass $\text{CoM}(G_k) = (x_k, y_k)$ to be the average of all robot locations $(x_i, y_i)$ and task locations $(x_j, y_j)$ for all $r_i, \in G_k^r$ and $t_j \in G_k^t$.

\begin{equation} \label{CoM}
\text{CoM}(G_k) = \frac{\sum_{r_i \in G_k^r} (x_i, y_i) + \sum_{t_j \in G_k^t} (x_j, y_j)}{|G_k^r|+ |G_k^t|} \end{equation}

Distances between $\text{CoM}(G_\alpha)$ and $\text{CoM}(G_\gamma)$ for all merge candidates  $G_\gamma \in M$ are converted into a probability mass function $P(G_\beta|G_\alpha)$ using a temperature-controlled softmax function \cite{sutton1998reinforcement}:
\begin{equation} \label{mergeDistribution} P_{G_\beta|G_\alpha}(G_\beta = G_\gamma | G_\alpha) = \frac{\exp(-d(G_\alpha, G_\gamma)/\tau)}{\sum_{G' \in M} \exp(-d(G_\alpha, G')/\tau)} \end{equation}
where $d(G_\alpha, G_\gamma)$ represents the Euclidean distance between $CoM(G_\alpha)$ and $CoM(G_\gamma)$, and $\tau$ is a tunable temperature parameter that controls the selectivity of the distribution. This probabilistic merging strategy favors combining nearby clusters while still exploring less obvious combinations.

\subsubsection{Heuristic Selection Approach}
The capability matching and spatial heuristics are combined into a single merge policy. First, the policy removes any merge candidates that do not satisfy the capability matching function (\ref{capMatch}):

\begin{equation}
    M' = \{G_\gamma \in M: m(G_\alpha, G_\gamma)=1\}
\end{equation}

Then merge candidate $G_\beta \in M'$ is chosen according to the spatial probability distribution defined in (\ref{mergeDistribution}).

\begin{equation}
    G_\beta \sim P_{G_\beta|G_\alpha}(G_\beta = G_\gamma|G_\alpha) \hspace{0.2em} \forall G_\gamma \in M'
\end{equation}

\subsection{Optimal Assignment within Clusters} \label{sec:clusterOptimization}

Given a cluster $G_k$ of $n$ robots and $m$ tasks, we seek an optimal assignment $A^*_k$ that maximizes the cluster's utility by assigning the cluster's robots $G_k^r$ to its tasks $G_k^t$. The optimization problem for each cluster closely mirrors the global problem in Equation (\ref{objective}), but with much smaller sets of robots and tasks:

\begin{equation}
\label{clusterObjective}
\begin{aligned}
   & \max_{A_k\in \mathcal{A}_k} \sum_{S_j \in A_k} U(S_j) \\
   & \text{subject to} \quad |S_j| \leq L, \quad \forall j: j \neq 0
\end{aligned}
\end{equation}
where $\mathcal{A}_k$ is the set of all possible assignments within the cluster.

Any method that finds the optimal assignment within the cluster can be used. We employ a branch and bound approach using integer partitions following the examples of \cite{prantare2020anytime} and \cite{rahwan2009anytime} where the search space is divided into partitions based on how many robots are assigned to each task, with upper bounds calculated for each partition. By exploring subspaces in descending order of these bounds and pruning those below our current best solution, the algorithm efficiently converges to the optimal assignment within each cluster.

\section{Analysis}
This section analyzes the complexity and convergence of Algorithm~\ref{alg:mainAlgorithm} and how cluster size constraints determine which suboptimal assignments can trap the algorithm.

\subsection{Computational Complexity}
There are $(\mu+1)^\nu$ possible assignments of $\nu$ robots to $\mu$ tasks plus an unassigned option. For a single cluster $G_k$, containing at most $L_r$ robots and $L_t$ tasks, the worst-case branch-and-bound complexity is $(L_t+1)^{L_r}$. The number of clusters is bounded above by $\nu+\mu$, as it cannot exceed the total number of robots and tasks, and bounded below by $\max(\nu/L_r, \mu/L_t)$ due to cluster size constraints. Both bounds scale linearly with $\nu$ and $\mu$, so the total number of clusters grows linearly with problem size. Consequently, the per-iteration time complexity of Algorithm~\ref{alg:mainAlgorithm} is
$$\mathcal{O}\big((L_t+1)^{L_r} \cdot (\nu + \mu)\big)$$
which is linear in $\nu$ and $\mu$. However, Algorithm~1 may require multiple iterations for convergence as discussed in the next section.

\subsection{Convergence}

We model the progression of Algorithm~\ref{alg:mainAlgorithm} as a finite Markov chain $\mathcal{M} = (\mathcal{A}, P)$, where $\mathcal{A}$ is the state space comprising all possible assignments, and $P(A,A')$ gives the probability of transitioning from assignment $A$ to $A'$ after one iteration of Algorithm~\ref{alg:mainAlgorithm}. Lemma \ref{nondecreasingLemma} shows that the utility of the assignment cannot decrease so $P(A,A') = 0$ when $\mathcal{U}(A') < \mathcal{U}(A)$. However, it could change between assignments of equal utility ($\mathcal{U}(A') = \mathcal{U}(A)$).

\begin{definition}\label{def:localtrapset}
A trap set $\Omega(u) \subseteq \mathcal{A}$ is a set of assignments that all produce the same utility value $u$ and cannot transition to any assignment outside the trap set.
\begin{equation}
\begin{aligned}
    \Omega(u) = \{A\in\mathcal{A}: \text{ }& \mathcal{U}(A)=u \text{ and } \\
    & P(A,A') = 0 \text{ for all } A' \notin \Omega(u)\}
\end{aligned}
\end{equation}
\end{definition}
Note that there is always at least one nonempty trap set associated with the set of globally optimal assignments $\mathcal{A}^*$ since any globally optimal assignment $A^* \in \mathcal{A}^*$ will produce the same optimal utility $u^*$ and by Lemma~\ref{nondecreasingLemma} cannot transition to an assignment that is suboptimal.

\begin{definition}
    The global trap set $\Omega = \bigcup_{u \in \mathbb{R}} \Omega(u)$ includes all assignments that trap the algorithm at any utility level $u \in \mathbb{R}$.
\end{definition}

\begin{definition} \label{def:reachable}
An assignment $A'$ is reachable from assignment $A$ (or equivalently $A$ can reach $A'$) if there exists a finite sequence of assignments $A = A^0, A^1, \ldots A^h = A'$ such that the probability of transitioning from $A^i$ to $A^{i+1}$ is non-zero for all $0 \leq i < h$.
\end{definition}

\begin{theorem} \label{convergenceThm}
Let $A^0, A^1, A^2, \ldots$ denote the sequence of assignments produced by Algorithm~\ref{alg:mainAlgorithm}. Let $A^N$ be the assignment obtained after $N$ iterations of Algorithm~\ref{alg:mainAlgorithm}. Then 
$$\lim_{N \to \infty} P(A^N \in \Omega) = 1.$$
\end{theorem}

\begin{proof}

First we show that every assignment $A^i \in \mathcal{A}$ can reach an assignment $\hat{A} \in \Omega$ in the global trap set. Let $\mathcal{A}^r$ be the set of assignments reachable from $A^i$, and let $u_{max} = \max_{A\in\mathcal{A}^r}\mathcal{U}(A)$ be the maximum utility produced by any assignment reachable from $A^i$. Define $\hat{\mathcal{A}}=\{ A \in \mathcal{A}^r: \mathcal{U}(A)=u_{max}\}$ to be the set of assignments reachable from $A$ that produce utility $u_{max}$. Consider any assignment $\hat{A} \in \hat{\mathcal{A}}$ and $A' \in {\mathcal{A}}$. We show $P(\hat{A},A') > 0 \implies A'\in \hat{\mathcal{A}}$, meaning any assignment in $\hat{\mathcal{A}}$ only transitions to other assignments in $\hat{\mathcal{A}}$.

First note that if $P(\hat{A},A') > 0$, then by Definition~\ref{def:reachable}, $A'$ must be reachable from $A^i$, meaning $A' \in \mathcal{A}^r$. By definition of $u_{\max}$, $\mathcal{U}(A') \leq u_{\max}$. By Lemma~\ref{nondecreasingLemma}, the utility cannot decrease, so $\mathcal{U}(A') = u_{max}$. It follows that:
\begin{align*}
P(\hat{A},A') > 0 \implies& A'\in A^r \text{ and } \mathcal{U}(A') = u_{max}\\
\implies& A' \in \hat{\mathcal{A}}.
\end{align*}
Therefore, $\hat{\mathcal{A}}$ satisfies Definition~\ref{def:localtrapset}, meaning $\hat{\mathcal{A}} \subseteq \Omega(u_{max})$. 

It follows that for any assignment $A^i \in \mathcal{A}$, there exists an assignment $\hat{A} \in \Omega$ in the global trap set that is reachable from $A^i$. By Definition~\ref{def:reachable}, this means there exists a finite sequence of assignments $A^i, A^1, \ldots, A^h = \hat{A}$ where each transition $A^i \to A^{i+1}$ has non-zero probability. Since $\mathcal{A}$ is finite, $h \leq |\mathcal{A}|$ for any such sequence. This implies that starting from any assignment $A^i \in \mathcal{A}$, there is a non-zero probability of entering the global trap set $\Omega$ within $|\mathcal{A}|$ iterations. Let $\epsilon$ denote the minimum probability of entering $\Omega$ within $|\mathcal{A}|$ steps, taken over all $A^i \in \mathcal{A}$. 

Now consider the algorithm after $N = k|\mathcal{A}|$ iterations for any positive integer $k$:
\begin{align*}
P(A^{k|\mathcal{A}|} \notin \Omega) &\leq (1-\epsilon)^{k} \\
P(A^{k|\mathcal{A}|} \in \Omega) & \geq 1 - (1-\epsilon)^{k}
\end{align*}
\vspace{-1ex}
$\lim_{N \rightarrow\infty} P(A^{N} \in \Omega) = \lim_{k \rightarrow\infty} P(A^{k|\mathcal{A}|} \in \Omega) = 1$.
\end{proof}



Because the global trap set $\Omega$ may contain local optima, the iterative clustering algorithm is not guaranteed to converge to the globally optimal solution. This is only to be expected since the underlying combinatorial optimization problem is NP hard. That said, techniques such as
restarting the algorithm or introducing random perturbations could be employed to modify our proposed algorithm to escape local optima. However, a detailed investigation of these strategies is left for future work.

\subsection{Characterizing Algorithm~\ref{alg:mainAlgorithm} Optimal Assignments}

We call any assignment $\hat{A} \in \Omega$ in the global trap set an Algorithm~\ref{alg:mainAlgorithm} optimal assignment, since by Definition~\ref{def:localtrapset}, it will never transition to an assignment with greater utility.

The cluster size constraints $L_r$ and $L_t$ determine which assignments are Algorithm~\ref{alg:mainAlgorithm} optimal. In the extreme case where $L_r = \nu$ and $L_t = \mu$, a single cluster would contain all robots and tasks. When optimized, this cluster would yield a globally optimal assignment. Tightening these constraints reduces computational complexity, but introduces more Algorithm~\ref{alg:mainAlgorithm} optimal assignments by limiting the algorithm's ability to explore larger reassignments.

\begin{definition} \label{def:reachableClusteringAlg}
A cluster formation algorithm is called universally reachable if for each feasible cluster (Definition~\ref{def:feasibleCluster}), the probability of forming that cluster is non-zero.
\end{definition}

Note that the uniform random clustering method is universally reachable, while the heuristic approach in Section~\ref{sec:mergeSelectionSection} is not, as it only merges clusters that fulfill specific capability matching criteria. Any algorithm that lacks universal reachability can be modified to achieve it by incorporating a small probability of random clustering.

\begin{theorem} \label{Thm_local_maxima}
Let $L_n = \min(L_r/L - 1, L_t - 1)$. If Algorithm~\ref{alg:mainAlgorithm} employs a universally reachable cluster formation algorithm, then for any assignment $\hat{A} \in \Omega$ in the global trap set, any utility improvement requires reassigning more than $L_n$ robots.
\end{theorem}

\begin{proof}
We show that any assignment $A = \{S_0, S_1, \ldots, S_\mu\}$ that can be improved by reassigning $x \leq L_n$ robots to reach a new assignment $A' = \{S'_0, S'_1, \ldots, S'_\mu\}$ with $\mathcal{U}(A) < \mathcal{U}(A')$ is not in the trap set. We demonstrate this by constructing a feasible cluster containing all robots and tasks necessary to perform a reassignment that increases utility.

Model the reassignment from $A$ to $A'$ as a graph $(N,E)$ where nodes $N = \{S_j \in A: S_j \neq S'_j\}$ represent teams modified by the reassignment, and each edge $e \in E$ represents a transfer of robots between two teams. We show that if graph $(N,E)$ is connected, then a feasible cluster can produce assignment $A'$, and if the graph is disconnected, then a smaller reassignment exists that increases utility.

\textbf{Case 1:} $(N,E)$ is a connected graph. Then to reassign $x$ robots, there are at most $|E| \leq x$ edges and at most $|N| \leq x+1$ modified teams. 

We construct a cluster $G_1$ containing all robots and tasks required to perform the reassignment from $A$ to $A'$. If $S_0$ (unassigned robots) is one of the modified teams ($S_0 \in N$), the cluster only needs to include the specific robots in $S_0$ being reassigned. However, for any assigned team $S_j \neq S_0$, constraint (\ref{eq:c}) requires including the entire team $S_j$ and its task $t_j$ to ensure robots are clustered with their currently assigned task. To determine maximum cluster size, we assume the worst case scenario where all robots are assigned to tasks: $S_0 \notin N$.

Let $G_1 = (G_1^r,G_1^t)$ where $G_1^r=\bigcup_{S_j \in N}S_j$ contains all robots in the modified teams and $G_1^t=\{t_j:S_j \in N, j \neq 0\}$ contains their associated tasks. This cluster enables the reassignment from $A$ to $A'$.

Since $|N| \leq x+1$, then $|G_1^r| \leq L(x+1)$ and $|G_1^t| \leq (x+1)$ where $L$ is the maximum team size. Therefore, cluster $G_1$ is feasible when:
\begin{align}
    L(x+1) &\leq L_r \implies x \leq \frac{L_r}{L} - 1\\
    x+1 &\leq L_t \implies x \leq L_t - 1
\end{align}
This is equivalent to $x \leq \min(L_r/L-1, L_t-1) = L_n$ which is the condition given in the theorem. 

\textbf{Case 2: } $\{N,E\}$ is not a connected graph. Then it can be decomposed into a set of at least two connected subgraphs $\{(N_1,E_1),\dots,(N_\ell,E_\ell)\}$ such that no robots are exchanged between teams in different subgraphs. 

Let $N' = \{S'_j \in A': S_j \in N\}$ and $\mathcal{U}(N) = \sum_{S_j \in N}\mathcal{U}(S_j)$. Since $\mathcal{U}(A) < \mathcal{U}(A')$ and only teams in $N$ are modified, it follows that:
\begin{equation}
\mathcal{U}(N_1) + \cdots + \mathcal{U}(N_\ell) < \mathcal{U}(N'_1) + \cdots + \mathcal{U}(N'_\ell)
\end{equation}

Therefore, for some $N_a \in \{N_1, \ldots, N_\ell\}$, $\mathcal{U}(N_a) < \mathcal{U}(N'_a)$. It follows that the utility of assignment $A$ can be increased by constructing a smaller cluster $G_2$ that includes only the teams $S_j \in N_a$. Since $(N_a,E_a)$ is a connected graph with fewer than $x$ edges, it follows from Case 1 that $G_2$ is feasible. 

In both cases, we have demonstrated the existence of a feasible cluster that, when optimized, will transition the assignment $A$ to one with higher utility. Given Definition~\ref{def:reachableClusteringAlg}, with each iteration, there is a nonzero probability of forming this cluster and making this transition. Therefore, $A$ is not in the global trap set.
\end{proof}
\section{Experimental Results}



\begin{figure}[t]  
    \centering
    
    \begin{subfigure}{\columnwidth}  
        \centering
        \includegraphics[width=.90\columnwidth]{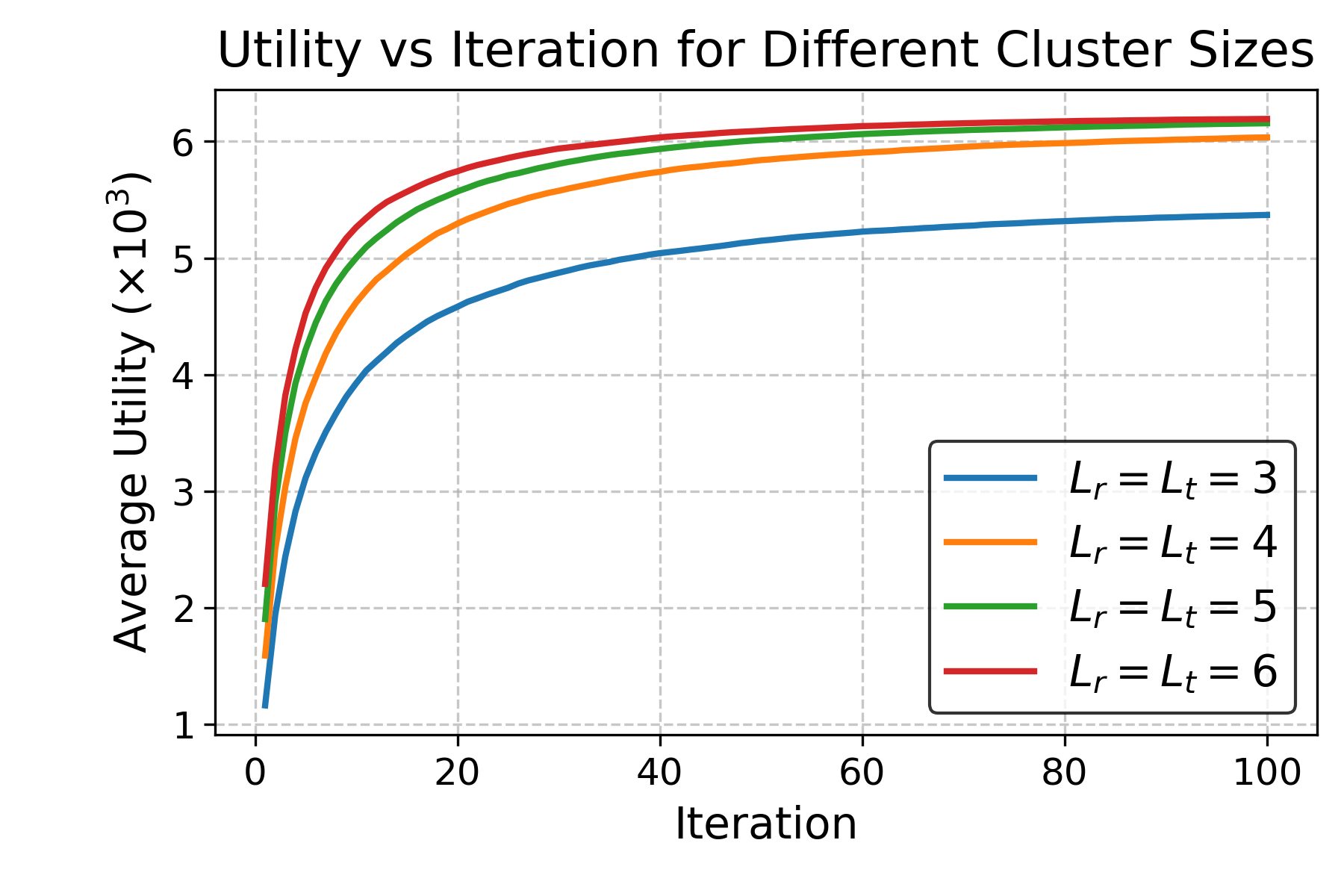}
    \end{subfigure}
    
    
    \begin{subfigure}{\columnwidth}  
        \centering
        \includegraphics[width=.90\columnwidth]{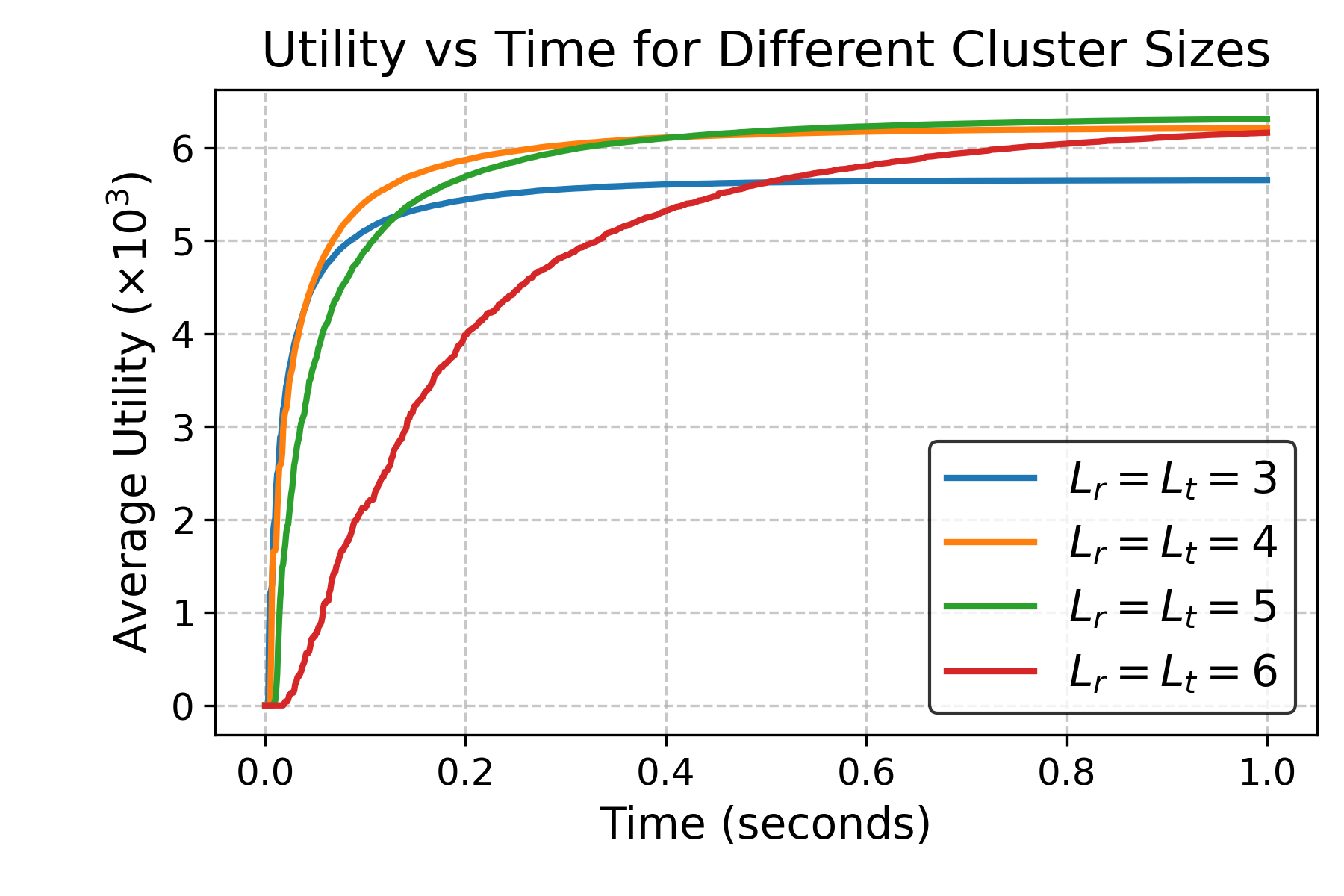}
    \end{subfigure}
    
    \caption{Utility produced by Algorithm~\ref{alg:mainAlgorithm} when employing the random clustering strategy, tested with various cluster size limits $L_r$ and $L_t$. Results are averaged over 100 tests with $\nu =100$ robots, $\mu = 50$ tasks, and $\kappa = 3$ robot types.}
    \label{fig:cluster_size_comparison}
    \vspace{-4mm}
\end{figure}

The iterative clustering algorithm is evaluated within a simulated $100 \times 100$ unit workspace, where $\nu$ robots and $\mu$ tasks are independently initialized at uniformly random positions. Robots are evenly partitioned into $\kappa=3$ types. Tasks are evenly divided into 10 task types, each associated with a benefit function $B_j(C)$ (Section~\ref{sec_util_modeling}) that maps the team composition $C$ to the utility for task $t_j$. The utility for each task depends on both the number and type of assigned robots and can exhibit linear returns, marginally decreasing returns, or marginally increasing returns (synergistic collaboration) as more robots are assigned. Each task’s team size is capped at $L=3$ robots, and the team-formation cost $J_j(S_j)$ is the cumulative distance traveled by assigned robots to the task location (Eq.~\ref{costFunction}). Unless otherwise noted, the iterative clustering procedure uses cluster-size limits $L_r=6$ (robots) and $L_t=3$ (tasks). Complete documentation, source code, and scripts to replicate all experiments and figures are available at https://github.com/AICPS/IterativeClusteringMRTA.


\subsection{Effect of Cluster Size} \label{sec:ClusterSizeComp}

The random clustering approach was used to evaluate the effects of different cluster sizes. As shown in Figure \ref{fig:cluster_size_comparison}, while smaller clusters require more iterations, they converge faster in time due to their lower computational cost per iteration. Cluster size limits $L_r$ and $L_t$ present a clear trade-off: larger clusters reduce convergence to suboptimal utilities and require fewer iterations to converge, but increase per-cluster computational complexity.

\subsection{Comparison to Optimal for Small Problems}

For small-scale problems, the clustering methods can be directly compared against optimal solutions. Table~\ref{tab:clustering_v_optimal} presents the performance comparison on 500 problem instances with $\nu = 10$ robots and $\mu = 5$ tasks. Both random and heuristic clustering methods achieve optimal or near-optimal utility, with no statistically significant difference observed between their results.

\begin{table}[h]
    \centering
    \caption{Clustering Methods Compared with Optimal}
    \label{tab:clustering_v_optimal}
    \begin{tabular}{lcc}
        \hline
        \\[-0.8em]
        Method & Exact Optimal Matches (\%) & Average Utility Ratio (\%) \\[0.5em]
        \hline
        \\[-0.8em]
        Random & 99.6 & 99.9 \\
        Heuristic & 99.0 & 99.9 \\[0.3em]
        \hline
    \end{tabular}
\end{table}
\vspace{-.4cm}
\subsection{Task Allocation Methods Comparison}

\begin{figure}[h]
\centering
\includegraphics[width=.85\columnwidth]{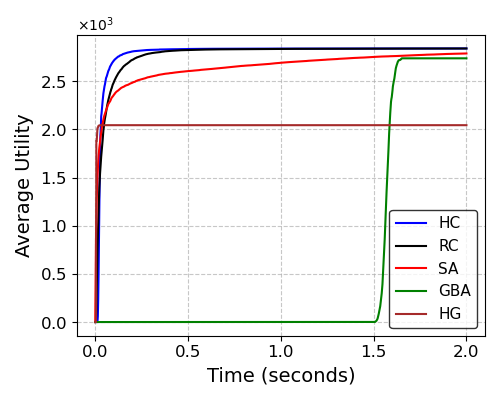}
\caption{Average algorithm performance over 100 tests for Heuristic Clustering (HC), Random Clustering (RC), Simulated Annealing (SA), Group-Based Auction (GBA), and Hedonic Game (HG) approaches.}
\label{fig:kappa3Results}
\end{figure}

For a larger problem with $\nu = 50$ robots and $\mu = 25$ tasks, heuristic and random iterative clustering methods were compared against a group-based auction, hedonic game, and simulated annealing. Overall, heuristic clustering outperformed the other methods, closely followed by random clustering. Figure \ref{fig:kappa3Results} shows the averaged results, and Table \ref{tab:algorithm_performance_3cap} reports mean utilities and standard deviations at 0.1, 0.5, and 2 seconds.

\begin{table}[h]
\caption{Algorithm performance: average utility, standard deviation, and maximum utility as percentages of the best solution found.}
\label{tab:algorithm_performance_3cap}
\centering
\renewcommand{\arraystretch}{1.3} 
\begin{tabular}{|l|cc|cc|ccc|}
\hline
\multirow{2}{*}{\textbf{Method}} & \multicolumn{2}{c|}{\textbf{0.1 seconds}} & \multicolumn{2}{c|}{\textbf{0.5 seconds}} & \multicolumn{3}{c|}{\textbf{2 seconds}} \\
\cline{2-8}
 & \textbf{Avg} & \textbf{$\sigma$} & \textbf{Avg} & \textbf{$\sigma$} & \textbf{Avg} & \textbf{$\sigma$} & \textbf{Max} \\
\hline
HC  & \textbf{95.2} & 2.6 & \textbf{99.4} & 0.5 & \textbf{99.6} & 0.4 & 100.0 \\
RC     & 87.4 & 3.4 & 99.1 & 0.6 & \textbf{99.6} & 0.4 & 100.0 \\
SA   & 82.7 & 2.7 & 91.4 & 1.5 & 97.8 & 0.9 & 99.5  \\
GBA   & 0.0  & 0.0 & 0.0  & 0.0 & 96.0 & 0.0 & 96.0  \\
HG          & 71.7 & 2.8 & 71.7 & 2.8 & 71.7 & 2.8 & 78.1  \\
\hline
\end{tabular}
\vspace{-5mm}
\end{table}

The group-based auction algorithm presented in \cite{Bai9781631} was adapted to assign teams of robots to tasks. In this version, task representatives evaluate all feasible teams and place bids based on each team’s average reward per robot. While it achieves similar utility to iterative clustering, it is far less computationally efficient since all feasible teams must be evaluated before the auction begins, whereas clustering considers only candidate teams within each cluster.

Hedonic games are a common heuristic for task allocation \cite{zhang2024coalition, GRAPE8439076}. We compare against a hedonic game where each robot iteratively joins the team that maximizes their marginal contribution, converging to a Nash equilibrium. Although computationally efficient, this method struggles on collaborative tasks requiring simultaneous multi-robot assignments, where individual reassignments provide no benefit.

The simulated annealing algorithm perturbs solutions by randomly replacing a single team while preserving other task assignments. Improvements are always accepted and worse solutions are accepted with decreasing probability over time. Simulated annealing converged more slowly than iterative clustering due to its randomized search approach, whereas iterative clustering employs a more directed search through optimal cluster assignments.

\subsection{Scalability}

\begin{figure}[h]
\centering
\includegraphics[width=.9\columnwidth]{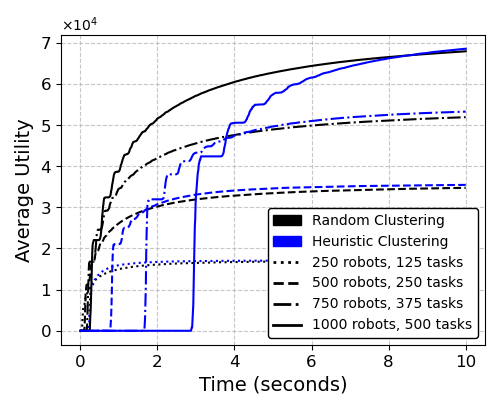}
\caption{Average algorithm performance on four different problem sizes.}
\label{fig:scalability_test}
\end{figure}

Algorithm~\ref{alg:mainAlgorithm}'s scalability was tested on four problem sizes, with results in Figure~\ref{fig:scalability_test} averaged over 100 trials. Random clustering scaled efficiently, while heuristic clustering degraded due to the cost of computing merge probabilities across many clusters. For the largest case (1000 robots, 500 tasks), the heuristic required about 3 seconds for the first iteration. Overall, the framework scales well, though larger problems demand alternative heuristics.

\vspace{-8mm}
\section{Conclusion}
This paper presented an iterative clustering algorithm for solving collaborative task allocation in multi-robot systems. The approach iteratively forms small clusters of robots and tasks, optimizing assignments within each cluster while ensuring robots remain clustered with their currently assigned tasks. This guarantees the utility is monotonically increasing with each iteration. Larger clusters avoid becoming trapped at suboptimal assignments, while smaller clusters offer faster convergence time. Experimental results validated the scalability, accuracy, and scalability of our approach compared to state of the art heuristics.

Future work will explore learning-based approaches to cluster generation, decentralized implementations of iterative clustering, and how clustering techniques can adapt to dynamic and uncertain environments.

\vspace{-2mm}

\bibliographystyle{IEEEtran}
\bibliography{references.bib}
\end{document}